\documentclass[sigconf]{acmart}
\AtBeginDocument{%
  }

\setcopyright{acmlicensed}
\copyrightyear{2026}
\acmYear{2026}
\setcopyright{cc}
\setcctype{by}
\acmConference[WWW '26]{Proceedings of the ACM Web Conference 2026}{April 13--17, 2026}{Dubai, United Arab Emirates}
\acmBooktitle{Proceedings of the ACM Web Conference 2026 (WWW '26), April 13--17, 2026, Dubai, United Arab Emirates}
\acmPrice{}
\acmDOI{10.1145/3774904.3792287}
\acmISBN{979-8-4007-2307-0/2026/04}

\usepackage[ruled,vlined,linesnumbered]{algorithm2e} % 加载宏包时指定注释格式

\usepackage{multirow}
\usepackage{pifont}
\usepackage{graphicx}
\usepackage{amsmath}
\usepackage{comment}
\usepackage{verbatim}
\usepackage{xcolor}
\usepackage{colortbl}

\usepackage{acronym}
\usepackage{fancyhdr}
\usepackage{balance}
\usepackage{booktabs}
\usepackage{arydshln} 
\begin{document}

% \title{MCoT-MVS: Multi-level Vision Selection by Multi-modal Chain-of-Thought Reasoning for Composed Image Retrieval}
\title[MCoT-MVS: Multi-level Vision Selection by Multi-modal Chain-of-Thought Reasoning \\ for Composed Image Retrieval]{MCoT-MVS: Multi-level Vision Selection by Multi-modal Chain-of-Thought Reasoning for Composed Image Retrieval}
%
% The "author" command and its associated commands are used to define
% the authors and their affiliations.
% Of note is the shared affiliation of the first two authors, and the
% "authornote" and "authornotemark" commands
% used to denote shared contribution to the research.
\author{Xuri Ge}
\authornote{Both authors contributed equally to this research.}
\orcid{1234-5678-9012}
\email{xuri.ge@sdu.edu.cn}
\affiliation{%
  \institution{Shandong University}
  \city{Jinan}
  \country{China}
}

\author{Chunhao Wang}
\authornotemark[1]
\affiliation{%
  \institution{Shandong University}
  \city{Qingdao}
  \country{China}}
\email{202415164@mail.sdu.edu.cn}

\author{Xindi Wang}
\affiliation{%
  \institution{Shandong University}
  \city{Jinan}
  \country{China}}
\email{xindi.wang@sdu.edu.cn}

\author{Zheyun Qin}
\affiliation{%
  \institution{Shandong University}
  \city{Qingdao}
  \country{China}}
\email{zheyunqin@gmail.com}

\author{Zhumin Chen}
\affiliation{%
  \institution{Shandong University}
  \city{Jinan}
  \country{China}}
\email{chenzhumin@sdu.edu.cn}

\author{Xin Xin}
\authornote{Corresponding author}
\affiliation{%
  \institution{Shandong University}
  \city{Qingdao}
  \country{China}}
\email{xinxin@sdu.edu.cn}

%
% By default, the full list of authors will be used in the page
% headers. Often, this list is too long, and will overlap
% other information printed in the page headers. This command allows
% the author to define a more concise list
% of authors' names for this purpose.
% \renewcommand{\shortauthors}{Ge et al.}
\renewcommand{\shortauthors}{Xuri Ge et al.}
%% No italics, no superscripts, not anonymous
%% Use footnote or author note to identify equal contribution and/or contact author info

\begin{abstract}
Composed Image Retrieval (CIR) aims to retrieve target images based on a reference image and modified texts. However, existing methods often struggle to extract the correct semantic cues from the reference image that best reflect the user's intent under textual modification prompts, resulting in interference from irrelevant visual noise. 
In this paper, we propose a novel Multi-level Vision Selection by Multi-modal Chain-of-Thought Reasoning (MCoT-MVS) for CIR, integrating attention-aware multi-level vision features guided by reasoning cues from a multi-modal large language model (MLLM). 
Specifically, we leverage an MLLM to perform chain-of-thought reasoning on the multimodal composed input, generating the retained, removed, and target-inferred texts. These textual cues subsequently guide two reference visual attention selection modules to selectively extract discriminative patch-level and instance-level semantics from the reference image.
Finally, to effectively fuse these multi-granular visual cues with the modified text and the imagined target description, we design a weighted hierarchical combination module to align the composed query with target images in a unified embedding space. 
Extensive experiments on two CIR benchmarks, namely CIRR and FashionIQ, demonstrate that our approach consistently outperforms existing methods and achieves new state-of-the-art performance. Code and trained models are publicly released at https://github.com/JJJJerry/WWW2026-MCoT-MVS.
\end{abstract}




\maketitle

% \acresetall
% \textbf{Relevance Statement:} This paper proposes an effective multimodal method for parsing user query intent for multimodal personalized retrieval, which more accurately captures the user's true intent. It aligns directly with the topic of "Search and retrieval-augmented AI" in the web track and can be deployed in practical web systems such as e-commerce platforms and search engines to enhance the functionality and effectiveness of multimodal retrieval.

\section{Introduction}

Composed Image Retrieval (CIR) \cite{vo2019composing} is a fundamental task in multimodal understanding, aiming to retrieve the target image corresponding to the user's intention based on the given reference image and modified texts. It requires a thorough understanding of the modification intent implicit in the modified texts to accurately perceive the relevant contents in the reference image. 
CIR task has garnered increasing attention in both academia and industry due to its broad real-world applications, such as personalized recommendations in e-commerce \cite{wu2024personalized,fu2025efficient} and interactive search engines \cite{xie2020modeling,luo2025imagescope}, etc. %For example, CIR supports interactive and conversational search in intelligent search engines, allowing users to iteratively refine their queries across multiple turns of dialogue.
The challenge of the CIR task is to correctly understand the user's intent from multimodal combined queries.
In particular, it is more challenging to integrate and align visual information from diverse reference image semantics related to the modification intent of the modification text to construct a joint representation that matches the target image.

 \begin{figure}[t] %%%%%%%%%%%%%%%%%fig1
    	\centering
         % \vspace{0.2em}
    	\includegraphics[width=1\linewidth]{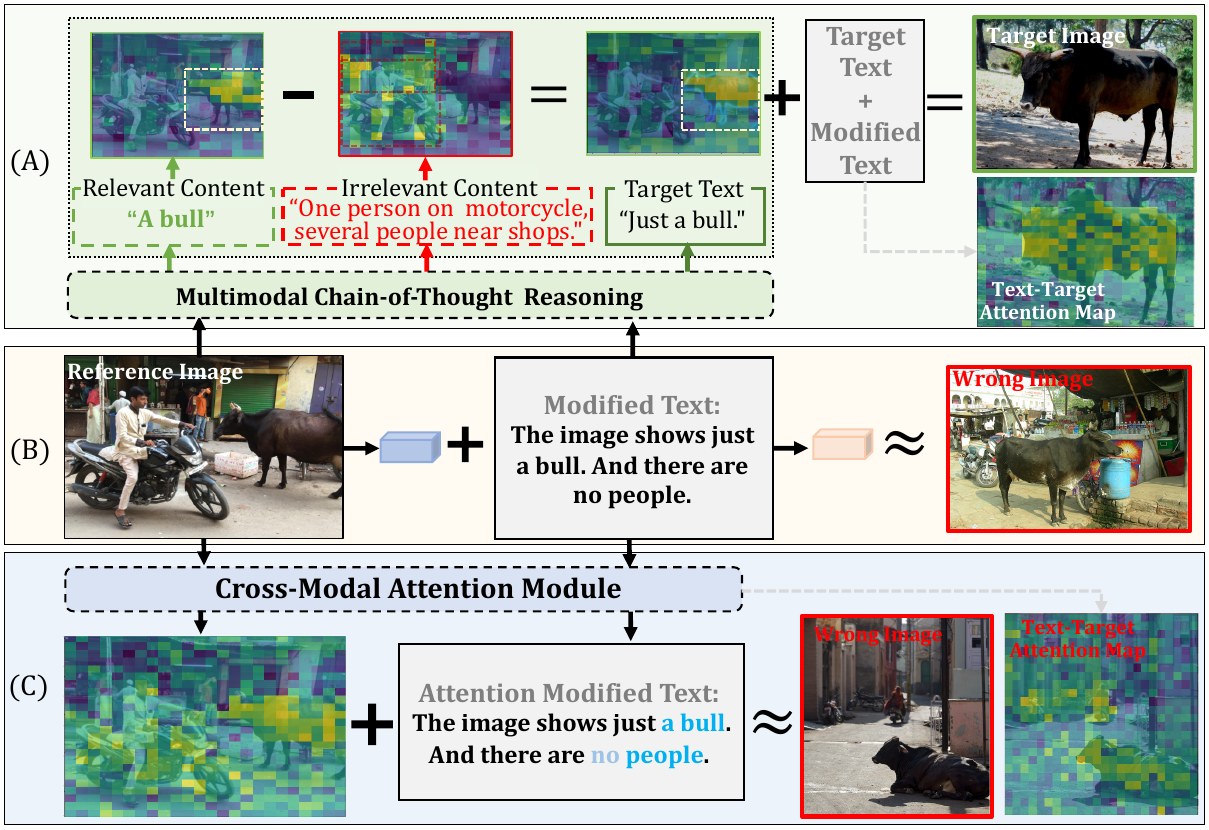}
    	\vspace{-2em}
        \caption{Compared with the traditional CIR model (B) and cross-modal attention-based CIR model (C), in this paper, we introduce multimodal chain-of-thought reasoning to explicitly extract the relevant and irrelevant contents for reference visual information selection (A).
        In this way, while effectively focusing on fine-grained useful references, we can significantly avoid the attention noise interference driven by visual-linguistic correlation as in method (C). 
        } %Compared with the original visual representation (), we introduce the contextual scene information from segmentation to highlight both object saliency and spatial saliency. Book bed mirror clothes table chair potted plant bookshelf floor
    	\label{fig:motivation}
    	\vspace{-1em}
    \end{figure} 
    
Early CIR methods \cite{kim2021dual,gudivada1995content,guan2022partially} primarily rely on joint embedding space modeling by fusing the global-level representations of the reference image and the modification text to match candidate target images. As shown in Figure \ref{fig:motivation}(B), these methods typically employ direct or low-level fusion strategies across modalities, lacking fine-grained semantic alignment. As a result, they struggle to accurately capture the visual modification contents based on the modified text. Especially when it comes to attribute detail changes and instance-level semantic replacement, the crude visual perception ability leads to visual noise in the combined query features.

To improve the representation ability of multimodal composed queries, recent methods \cite{chen2020image,wen2021comprehensive,tian2023fashion} incorporate cross-modal attention mechanisms to establish local or global semantic associations between reference images and modification texts, as illustrated in Figure \ref{fig:motivation}(C). Attention-based fusion can enhance both the intent-aware representation of composed queries and their semantic consistency with potential target images. For example, \citet{xu2023multi} introduces multiscale attention-aware correspondences between reference images and modified texts to enhance composed-query representations, while \citet{tian2023fashion} employs additive attention to capture the intended semantics from multimodal queries.
Despite the certain effectiveness of attention mechanisms in improving joint multimodal representations, existing methods face clear limitations in interpreting complex composed-query intents and selecting visual regions aligned with them. %Specifically, these methods are largely driven by superficial similarity between the reference image and the modification text, lacking explicit semantic reasoning. 
Specifically, these methods typically rely on shallow appearance semantic correlations between salient objects in the reference image and the modification text, lacking explicit semantic reasoning. As shown in Figure \ref{fig:motivation}(C), they struggle to distinguish implicit visual operations in the modification, such as retaining ``a bull'' or removing ``all people, motorcycle, and shop", etc.  
This results in reference image content (e.g., ``a people on a motorcycle''), which should be removed, being incorrectly included in the combined query due to a spurious semantic similarity attention mechanism. This undermines the retrieval intent and degrades retrieval performance.

To deal with the above issue, we argue that an intuitive motivation shown in Figure \ref{fig:motivation}(A) is to decompose the user's composed query intent and integrate the correct reference visual semantics aligned with the intended modification, thereby reducing the ambiguity of the composed query. 
On the one hand, the rapid progress of Large Language Models (LLMs) offers powerful capabilities for intent reasoning. The latest zero-shot CIR methods \cite{sun2025cotmr,sun2025leveraging} also introduce to reason about user modification intentions based on LLMs. However, these studies primarily rely on textual cues without effective interaction with the visual modality, resulting in composed queries still being polluted by irrelevant visual features.
On the other hand, effectively distinguishing the useful reference image elements is crucial for enhancing visual grounding in composed queries. The representative solutions are in two folds. One is to process the global grid features (i.e., patch) \cite{kim2021dual,wen2024simple}, and the other is instance-level representations based on object detection \cite{hosseinzadeh2020composed}. While attention-aware visual interactions are then performed separately on these features, their selection is often still biased by visual-linguistic similarity, lacking explicit intent grounding. %Then visual feature interactions are performed on them separately to enhance the visual representation of user intent. However, these approaches are also disturbed by the visual-linguistic similarity of attention implicitly.

\begin{figure*}[t] %%%%%%%%%%%%%%%%%fig2
    	\centering
        % \vspace{-0.5em}    
    	\includegraphics[width=1\linewidth]{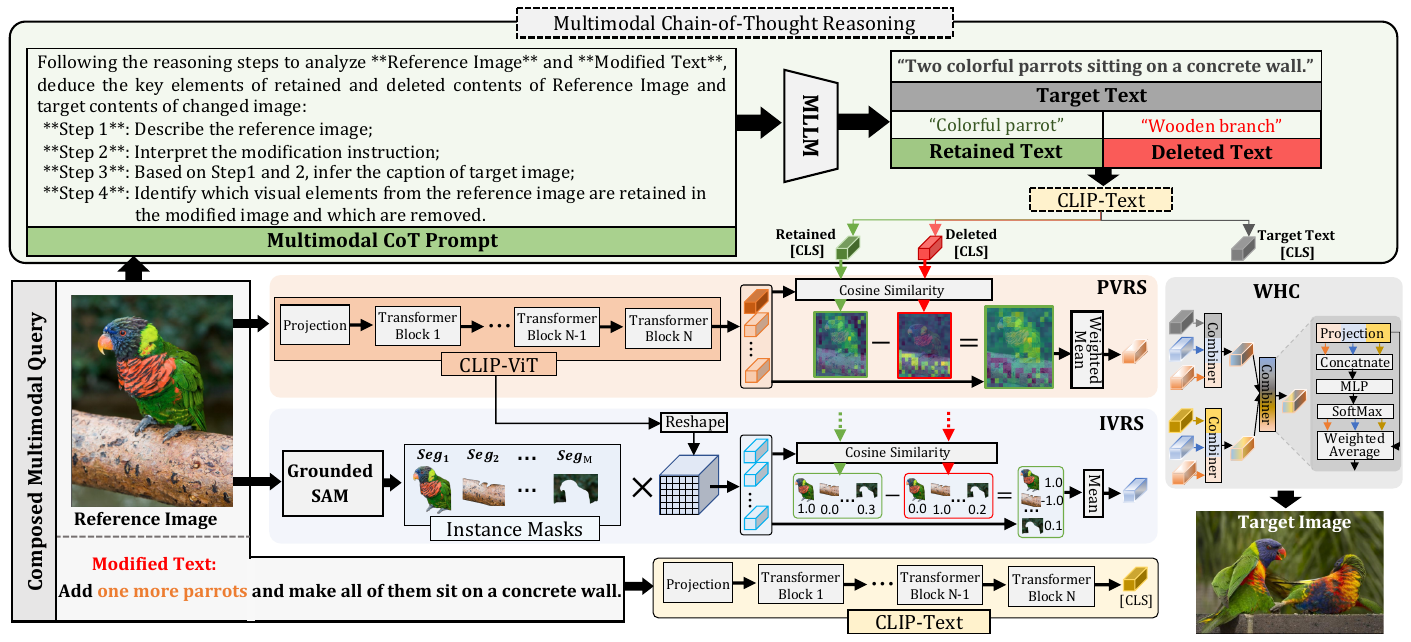}
    	\vspace{-2em}
     \caption{Illustration of our proposed MCoT-MVS. It primarily consists of the Multimodal Chain-of-Thought (CoT) Reasoning module, Patch-level Visual Reference Selection (PVRS), the Instance-level Visual Reference  Selection module (IVRS), and a Weighted Hierarchical Combination (WHC). The multimodal CoT reasoning module first reasons the composed query into retained and deleted contents, as well as the potential target context, based on a pre-trained MLLM. Then, the useful patch-level and instance-level reference visual representations are semantically selected by the explicitly reasoned retained and deleted texts to retain the correct user intent and to remove visual noise. Finally, the WHC fuses the selected representations with multiple target texts into an attention-aware query, aligning with the target image. 
     }
    	\label{Framework}    
     	\vspace{-1em}
    \end{figure*}

Therefore, integrating intent decomposition with effective reference visual feature selection is a natural choice for explicitly modeling composed-query intent and enhancing multimodal representations.
As manifested in Figure \ref{fig:motivation}(A), reasoning explicitly about both relevant and irrelevant elements, along with the target context, enables more precise localization of visual intention and avoids distraction from irrelevant content, compared to Figure \ref{fig:motivation}(C). 

In this paper, we propose a novel CIR framework, \textbf{MCoT-MVS} (Multi-level Vision Selection via Multi-modal Chain-of-Thought Reasoning). The framework first performs chained reasoning on multimodal composed queries using a multimodal large-scale language model (MLLM) to generate structured linguistic cues that explicitly mark semantic content to be retained, removed, and user-intended goals.  Subsequently, we design two vision selection modules to operate on both patch-level and instance-level features of the reference image, enabling multi-granularity extraction of semantically aligned visual cues. Finally, a weighted hierarchical combination (WHC) module fuses the selected visual features with the modification and referred target text to further clarify user intent, aligning the final composed representation with the correct target image in a shared embedding space.

The contributions of this paper are as follows:
 \begin{itemize} 
     \item We propose a novel MCoT-MVS with a multi-modal chain-of-thought reasoning for explicit multi-level reference visual element selection with hierarchical combination. It effectively reduces irrelevant multimodal noise of queries in composed image retrieval.
     \item We design multi-level reference vision selection modules, including a Patch-level Visual Reference Selection (PVRS) and an Instance-level Visual Reference Selection (IVRS), which decompose and select reference image contents at both fine-grained and high-level semantics. This improves the accuracy of the reference visual intent. 
    \item We introduce a Weighted Hierarchical Combination (WHC) to effectively fuse selected visual cues with modified text and reasoned user-intended text in a unified embedding space, enabling precise alignment with target images.
\end{itemize}
The proposed MCoT-MVS is verified to achieve the state-of-the-art (SOTA) retrieval performance on two standard composed image retrieval benchmarks, \textit{i.e.} CIRR and FashionIQ. %In particular, we achieve significant margins over the best method on CIRR (natural scene dataset), such as 55.3\% vs. other 53.4\% in Recall@1, because we have a clear advantage of inference denoising in more complex scenarios.

\section{Related Works}
%差不多一栏，半页即可。
\textbf{CIR with Direct Composed Query Fusion.} Early approaches to Composed Image Retrieval (CIR) \cite{chen2020learning,kim2021dual,wen2024simple} focus on directly combining the reference image and the modified text to construct a joint query representation. These methods typically adopt either global feature concatenation or local feature fusion. For example, \citet{chen2020learning} proposed a unified joint visual semantic matching model that learns image-text directly compositional embeddings by jointly associating visual and textual modalities in a shared discriminative embedding space via compositional losses.
\citet{kim2021dual} introduced a dual composition network to learning a joint embedding space by constructing two cyclic directional mappings for global embedding of triplets (reference image, text query, target image) via both composition and correction.
While these methods offer a straightforward way to integrate modalities, they often struggle to disentangle which aspects of the composed query should be retained, removed, or replaced based on the modification intent.

\textbf{CIR with Attention-based Composed Query Fusion.}
To improve the semantic alignment between reference images and modified text, attention-based CIR models \cite{delmas2022artemis,wen2021comprehensive,tian2023fashion} have been developed. These approaches typically learn cross-modal interactions via self- and cross-attention modules between the multimodal composed queries. For instance, \citet{delmas2022artemis} ensembled multiple target-query similarity scores by exploring the attention mechanisms between the reference image and modified text, as well as target images and modified text.  \cite{baldrati2023composed} proposed a simple soft attention-aware combiner module to compose the multimodal queries from the pre-trained foundation vision-language models, achieving the effective fusion of multimodal combination features from the pre-trained model. However, they usually rely on shallow similarity patterns and lack deeper semantic reasoning capabilities, which limits their ability to understand complex modification intents.

\textbf{CIR with LLM-guided Intent Reasoning and Understanding.}
Recent advances leverage Large Language Models (LLMs) and Multi-modal Large Language Models (MLLMs) to enhance CIR with high-level reasoning. Some works \cite{karthik2023vision} introduce intent rewriting, where the composed query is reformulated into a target description using LLMs. It has shown strong generalization in zero-shot CIR settings. 
Others, such as CIR-LVLM \cite{sun2025leveraging,sun2025cotmr}, fine-tuned VLM models to align image-text features through latent intent inference. These methods benefit from external knowledge and zero-shot generalization but typically treat the reference image as a whole, without fine-grained control over different levels of visual content. Our work differs by explicitly modeling multi-level vision selection (patch-level and instance-level) guided by MLLM-based multi-step reasoning, enabling interpretable and controllable fusion of reference semantics.

\section{The Proposed Method -- MCoT-MVS}
The overview of MCoT-MVS is illustrated in Figure \ref{Framework}. The innovation of MCoT-MVS lies in improving the consistency between multimodal composed queries and target images by effectively decomposing composed query intent (Multimodal CoT Reasoning), selecting useful multi-level query references (PVRS and IVRS) and reconstructing multimodal combined query representation (WHC). 

\subsection{Preliminary}
% This paper proposes a novel MCoT-MVS method to solve the task of target image $I^T$ retrieval by the given multimodal composed query includes a reference image $(I^Q)$ and modification texts.
We propose a novel MCoT-MVS approach for CIR, aiming to effectively retrieve the target image $I^T$ given a multimodal composed query consisting of a reference image $I^R$ and the corresponding modification texts $T$. During training stage, given the triplet training set $\mathcal{G}=\{(I^R_i,T_i,I^T_i)\}_{i=1}^\Omega$, a joint query-target embedding space is optimized by narrowing the embedding distance between the representation of the multimodal query ($\mathcal{F}(I^R_i,T_i)$) and the corresponding representation of target image ($\mathcal{H}(I^T_i)$). In this paper, we employ a widely used loss function in the CIR research community, i.e., batch-based classification loss \cite{vo2019composing}, formulated as,
\begin{equation}
    % \begin{small}
    %     % \begin{gather}
    %         % \begin{aligned}
            \mathcal{L} = \frac{1}{B}\sum_{i=1}^{B}-{\rm log} \left(\frac{{\rm Exp} \{{\rm Cos}(\mathcal{F}(I^R_i,T_i), \mathcal{H}(I^Q_i)) /\tau \} }{\sum_{j=1}^{B} {\rm Exp} \{{\rm Cos}(\mathcal{F}(I^R_i,T_i), \mathcal{H}(I^Q_j))/\tau\} }\right),
            \label{eq:loss}
            % \end{gather}
        % \end{small}
\end{equation}
where $B$ is the batch size, $\tau$ is the temperature coefficient and ${\rm Cos}(\cdot,\cdot)$ denotes the cosine similarity function. Our final aim is to optimise all parameters of $\mathcal{F}$ and $\mathcal{H}$, where $\mathcal{F}()$ for the composed query encoding contains the main contributions of our proposed MCoT-MVS shown in Figure \ref{Framework}. Algorithm \ref{alg:mcot_mvs_core} details the training procedure of MCoT-MVS.
%is as close as possible to the corresponding target image.

\subsection{Multimodal Chain-of-Thought Reasoning}
To fully understand the user's composed query intention, especially the heterogeneous and complex expression of multimodal queries, we first propose a multimodal chain-of-thought (CoT) reasoning method based on a pre-trained MLLM to effectively decompose the multimodal query intention and explicitly separate the query intention. Similar to \cite{sun2025cotmr}, as shown in Figure \ref{Framework}, we design a multi-step CoT reasoning prompt for the pretrained MLLM, including reference image understanding, modified text understanding, content inference of the corresponding target image, and decomposition of the visual elements to be retained or deleted. Different to other CoT reasoning CIR models \cite{sun2025cotmr}, we use a unified multimodal CoT reasoning prompt to decompose and reorganise the query intentions, which maintains both high efficiency and effectiveness. After that, the multimodal composed query is decomposed into the retained text $RT$ and the deleted text $DT$ and is reconstructed into a potential target text $TT$.
\begin{equation}
            % \begin{aligned}
            {RT,DT,TT}={\rm MLLM}(I^R,T;{\rm CoT\ Prompt}),
\end{equation}
Then, we employ the text encoder of CLIP~\cite{radford2021learning} to extract the representations of these reasoning results as follows: %$\{R^{RT},R^{DT},R^{TT}\}={\rm CLIP}$-${\rm Text}(RT/DT/TT)$. 
\begin{equation}
            % \begin{aligned}
           \{R^{RT},R^{DT},R^{TT}\}={\rm CLIP \mbox{-} Text}(RT/DT/TT),
\end{equation}
where $\{R^{RT},R^{DT}\}$ will be used for the multi-level visual reference selection and 
$R^{TT}$ for final query combination.

\subsection{Multi-level Visual Reference Selection}
Unlike the explicit semantics of the modified text, selecting relevant visual elements from the reference image in CIR is more challenging, requiring both accurate intent understanding and effective filtering of irrelevant content to reduce ambiguity.
To enhance the effectiveness of reference image representation, we propose patch-level and instance-level visual reference selection modules, termed PVRS and IVRS. Guided by the preceding reasoning results, i.e. $R^{RT}$ for retained semantic and $R^{DT}$ for deleted semantic, we explicitly reconstruct attention for both patch-level and instance-level visual reference features based on inferred user intent. This mechanism emphasizes the retained visual elements while suppressing those related to significant modifications or replacements, thereby improving the reference image's relevance to the composed query.

\begin{algorithm}[t]
\caption{Pseudo code of MCoT-MVS.}
\label{alg:mcot_mvs_core}
\SetAlgoLined
\KwIn{Training set $\mathcal{G}=\{(I^R_i,T_i,I^T_i)\}_{i=1}^\Omega$, pre-trained MLLM, pre-trained Grounded-SAM, pre-trained CLIP, batch size $B$, epochs $E$} %, temperature $\tau$}
\KwOut{Trained MCoT-MVS model}

Initialize CLIP-ViT, CLIP-Text, PVRS, IVRS, WHC modules\;

\For{$epoch = 1$ \KwTo $E$}{
    Shuffle $\mathcal{G}$; split into batches of size $B$\;
    \For{each batch}{
        % CoT Reasoning
        % \textcolor{gray}{\tcp{Step 1: Multi-modal CoT Reasoning}}
        \For{each composed query $(I^R_i,T_i)$}{
            \textcolor{gray}{{// Step 1: Multi-modal CoT Reasoning}}\\
            % Generate retained/deleted/potential target texts:\\
            $\{RT_i, DT_i, TT_i\}$ $\gets$ MLLM($I_i^R,T_i$, $CoT Prompt$)\;
            % Encode via CLIP-Text: $R^{RT}, R^{DT}, R^{TT}$\;
            $\{R_i^{RT},R_i^{DT},R_i^{TT}\}$ $\gets$ ${\rm CLIP \mbox{-} Text}(RT_i/DT_i/TT_i)$\;
        % }
         \textcolor{gray}{{// Step 2: Multi-level Visual Reference Selection}} \\
            $\hat{V}_i^P$ $\gets$ PVRS (Eq.(\ref{eq.4}-\ref{eq.6}))\;
            $\hat{V}_i^I$ $\gets$ IVRS (Eq.(\ref{eq.7}-\ref{eq.9})) \;
            % Selected instance features $\hat{V}_i^I$ via Eq.(\ref{eq.7}-\ref{eq.9}) in IVRS\;
        % }
        % WHC: Unified Query Representation
        \textcolor{gray}{{// Step 3: Weighted Hierarchical Combination}} \\
        % \For{each sample}{
            Encode modification text: $S_i^{MT}$=CLIP-Text($T_i$)\;
            $Q_i^M = \text{Combiner}(\hat{V}_i^P, \hat{V}_i^I, S_i^{MT})$\;
            $Q_i^T = \text{Combiner}(\hat{V}_i^P, \hat{V}_i^I, R_i^{TT})$\;
            $\mathcal{F}_i = \text{Combiner}(Q_i^M, Q_i^T)$\;
        }
        
        % Target features and loss
        Extract target image: $\mathcal{H}_i = \text{CLIP-ViT}(I^T_i)$\;
        % $\mathcal{L} = -\frac{1}{B}\sum_i \log \frac{\exp(\text{Cos}(\mathcal{F}_i, \mathcal{H}_i)/\tau)}{\sum_j \exp(\text{Cos}(\mathcal{F}_i, \mathcal{H}_j)/\tau)}$\;
        
        % Backpropagation
        Update parameter of CLIPs, PVRS, IVRS, WHC modules according to Loss function $\mathcal{L}$ in Eq.\ref{eq:loss}.\;
    }
}

% \Return Trained MCoT-MVS model\;
\end{algorithm}

    For the fine-grained PVRS module, we first extract the patch-level representation of the reference image $I^R$ using the CLIP-ViT encoder~\cite{radford2021learning}, denoted as $V^P=\{[CLS],v^P_1,...,v^P_N\}$, where $v^P_i$ is the feature of i-th visual patch. Then, we leverage the previously generated reasoning representations $R^{RT}$ and $R^{DT}$ to compute the cosine similarity with each visual patch, serving as attention weights for identifying retained and deleted areas. 
        \begin{equation} \label{eq.4}
            % \begin{aligned}
           \alpha_i^{P_{+}}={\rm Cos}(v^P_i , R^{RT}) = \frac{v^P_i \cdot R^{RT}}{\Vert v^P_i \Vert \Vert R^{RT} \Vert}, 
        \end{equation}
        \begin{equation}
           \alpha_i^{P_{-}}={\rm Cos}(v^P_i , R^{DT}) = \frac{v^P_i \cdot R^{DT}}{\Vert v^P_i \Vert \Vert R^{DT} \Vert},
        \end{equation}
    The final refined patch-level reference visual representation $\hat{V}^P$ is obtained by contrastively re-weighting the original feature with the retain/delete attentions, formulated as,
    % : $\hat{v}_i^P=\alpha_i^{retain} \cdot v_i^P - \alpha_i^{delete} \cdot v_i^P$
\begin{equation} \label{eq.6}
            % \begin{aligned}
           \hat{V}^P=\left(\frac{1}{N}\sum_{i=1}^{N}(\alpha_i^{P_{+}}- \alpha_i^{P_{-}}) v_i^P+V^P_{[CLS]} \right)/2,
\end{equation}
    where $N$ is the number of patches of the reference image.
    This emphasises reference visual patches aligned with retained semantics and suppresses visual noise, enhancing the fine-grained semantic relevance in CIR.

    Beyond patch-level feature selection, we further propose a high-level IVRS module, which captures more complete instance semantics and shape-aware information. Specifically, we first utilize the pre-trained Grounded SAM model~\cite{ren2024grounded} to segment the reference image $I^R$ into $M$ potential instance areas, denoted as ${Seg_i}_{i=1}^M$. These instance masks are then used to extract the visual features of the reference objects $O^R = {o^R_1, ..., o^R_M}$ via masked average pooling over the CLIP-ViT encoder~\cite{radford2021learning} feature maps. Subsequently, we compute the cosine similarity between each instance feature $o^R_i$ and the previously generated reasoning representations for "Retain" and "Delete" semantics, $R^{RT}$ and $R^{DT}$, to obtain the corresponding attention weights $\alpha_i^{RT}$ and $\alpha_i^{DT}$:
\begin{equation}\label{eq.7}
           \alpha_i^{I_{+}}= \frac{v^I_i \cdot R^{RT}}{\Vert v^I_i \Vert \Vert R^{RT} \Vert},
           \alpha_i^{I_{-}} = \frac{v^I_i \cdot R^{DT}}{\Vert v^I_i \Vert \Vert R^{DT} \Vert},
\end{equation}
    We then calculate the instance attention weights $\bar{\alpha}_i^{I}$ by difference the retained weights and deleted weights after min-max normalization. Finally, the refined instance-level reference visual representation $\hat{V}^I$ is aggregated by weighted mean:
            \begin{equation}
           \bar{\alpha}_i^{I_*} = \frac{\alpha_i^{I_{*}}-{\rm min}(\{\alpha^{I_{*}}\})}{{\rm max}(\{\alpha^{I_{*}}\})-{\rm min}(\{\alpha^{I_{*}}\})} ,
           \end{equation}
           \begin{equation}\label{eq.9}
           \hat{V}^I=\frac{1}{M} \sum_{i=1}^{M}( \bar{\alpha}_i^{I_{+}} - \bar{\alpha}_i^{I_{-}}) v_i^I  ,
           \end{equation}
    where $I_*$ is ${I_+\ {\rm or}\ I_-}$.

\subsection{Weighted Hierarchical Combination}
After multi-level visual selection guided by explicit reasoning cues from multimodal CoT reasoning, we obtain multiple multimodal query representation, including the patch-level visual reference feature $\hat{V}^P$, the instance visual reference feature $\hat{V}^I$, the modified textual feature $S^{MT}$ extracted from the CLIP-Text encoder, and the reasoned target textual feature $R^{TT}$. To effectively integrate these multimodal representations and accurately reflect user intent, we propose a Weighted Hierarchical Combination (WHC) module. This module performs hierarchical and adaptive fusion of the different query features under alignment supervision with the target image. 

Specifically, we first fuse the modified feature with multi-level visual features using adaptive weights, allowing features that more clearly reflect the user's intent to be emphasized. In parallel, we integrate the inferred target semantics with visual references to mitigate noise from incorrect reasoning. Inspired by \cite{baldrati2023composed}, we revise the $Combiner$ (see Figure \ref{Framework}) with triplet inputs (Corresponding textual representation, PVRS's visual representation and IVRS's visual representation). It consists of a concatenation operation with projecting, an MLP attention computation, and a weighted aggregation.
We take the Modification $Combiner$ process as an example, formulated as,
            \begin{equation}
           % Q = {\rm Combiner} \left( {\rm Combiner}(\hat{V}^P,\hat{V}^I, S^{MT}), {\rm Combiner}(\hat{V}^P,\hat{V}^I, S^{MT}) \right),\\
           V^{cat} = [W_1 \hat{V}^P : W_2 \hat{V}^I: W_3 S^{MT}],
           \end{equation}
           \begin{equation}
           \{\beta_1,\beta_2,\beta_3\} = {\rm SoftMax}( {\rm MLP}( V^{cat})),
           \end{equation}
           \begin{equation}
           Q^M=  \beta_1* \hat{V}^P + \beta_2 * \hat{V}^I + \beta_3 * S^{MT} +MLP(V^{cat}).
           \end{equation}
    A final $Combiner$ fusion combines both streams $Q^M$ and $Q^T$ into a unified query representation $Q$. 
    
    Finally, the unified query representation $Q$ from our proposed MCoT-MVS (i.e. $\mathcal{F}()$) is used to align with the target image feature (i.e. $\mathcal{H}(I^T)$).

% CIRR
\begin{table}[t]
\caption{Performance comparison on CIRR from the open-world scene. The best and the second-best scores are marked in bold and underlined fonts, respectively.}
\label{tab:tab1_CIRR}
\vspace{-1em}
% \fontsize{9}{9.5}\selectfont
\renewcommand\tabcolsep{3pt}
\begin{tabular}{llllll}
\hline
\multicolumn{1}{c|}{Method} & \multicolumn{1}{c}{R@1} & \multicolumn{1}{c}{R@5} & \multicolumn{1}{c}{R@10} & \multicolumn{1}{c|}{R@50} & \multicolumn{1}{c}{Avg} \\ \hline
\multicolumn{6}{c}{\cellcolor[HTML]{EFEFEF}\textit{Traditional Model-Based Methods}} \\
\multicolumn{1}{l|}{TIRG \cite{vo2019composing} $_{(CVPR'19)}$} & 14.61 & 48.37 & 64.08 & \multicolumn{1}{l|}{90.03} & 54.27 \\
\multicolumn{1}{l|}{CIRPLANT \cite{liu2021image} $_{(ICCV'19)}$} & 19.55 & 52.55 & 68.39 & \multicolumn{1}{l|}{92.38} & 58.22 \\
\multicolumn{1}{l|}{ARTEMIS \cite{delmas2022artemis} $_{(ICLR'22)}$} & 16.96 & 46.10 & 61.31 & \multicolumn{1}{l|}{87.73} & 53.03 \\
\multicolumn{6}{c}{\cellcolor[HTML]{EFEFEF}\textit{VLP Model-Based Methods}} \\
\multicolumn{1}{l|}{CLIP4CIR \cite{baldrati2023composed} $_{(TOMM'23)}$} & 38.53 & 69.98 & 81.86 & \multicolumn{1}{l|}{95.93} & 71.58 \\
\multicolumn{1}{l|}{TGCIR \cite{wen2023target} $_{(ACM\ MM'23)}$} & 45.25 & 78.29 & 87.16 & \multicolumn{1}{l|}{97.30} & 77.00 \\
\multicolumn{1}{l|}{SADN \cite{wang2024semantic} $_{(ACM\ MM'24)}$} & 44.27 & 78.10 & 87.71 & \multicolumn{1}{l|}{97.89} & 76.99 \\
\multicolumn{1}{l|}{DQU-CIR \cite{wen2024simple} $_{(SIGIR'24)}$} & 46.22 & 78.17 & 87.64 & \multicolumn{1}{l|}{97.81} & 77.46 \\
\multicolumn{1}{l|}{CaLa \cite{jiang2024cala} $_{(SIGIR'24)}$} & 49.11 & 81.21 & 89.59 & \multicolumn{1}{l|}{98.00} & 79.48 \\
\multicolumn{1}{l|}{SSN \cite{yang2024decomposing} $_{(AAAI'24)}$} & 43.91 & 77.25 & 86.48 & \multicolumn{1}{l|}{97.45} & 76.27 \\
\multicolumn{1}{l|}{CASE \cite{levy2024data} $_{(AAAI'24)}$} & 49.35 & 80.02 & 88.75 & \multicolumn{1}{l|}{97.47} & 78.90 \\
\multicolumn{1}{l|}{CoVRBLIP \cite{ventura2024covr} $_{(AAAI'24)}$} & 49.69 & 78.60 & 86.77 & \multicolumn{1}{l|}{94.31} & 77.34 \\
\multicolumn{1}{l|}{SPRC \cite{bai2023sentence} $_{(ICLR'24)}$} & 51.96 & 82.12 & 89.74 & \multicolumn{1}{l|}{97.18} & 80.25 \\
\multicolumn{1}{l|}{ENCODER \cite{li2025encoder} $_{(AAAI'25)}$} & 46.10 & 77.98 & 87.16 & \multicolumn{1}{l|}{97.64} & 77.22 \\
\multicolumn{1}{l|}{CIRLVLM \cite{sun2025leveraging} $_{(AAAI'25)}$} & \underline{53.64} & 83.76 & 90.60 & \multicolumn{1}{l|}{97.93} & 81.48 \\
\multicolumn{1}{l|}{CCIN \cite{tian2025ccin}  $_{(CVPR'25)}$} & 53.41 & \underline{84.05} & \underline{91.17} & \multicolumn{1}{l|}{\underline{98.00}} & \underline{81.66} \\ \hline
\multicolumn{1}{l|}{\textbf{MCoT-MVS$_{\rm (Ours)}$}} & \textbf{55.33} & \textbf{84.75} & \textbf{91.45} & \multicolumn{1}{l|}{\textbf{98.55}} & \textbf{82.52} \\ \hline
\end{tabular}
\end{table}

%fiq
\begin{table*}[t]
\caption{Performance comparison on FashionIQ from fashion products. The best and the second-best scores are marked in bold and underlined fonts, respectively.}
\vspace{-1.2em}
% \fontsize{9}{9}\selectfont
\renewcommand\tabcolsep{12pt}
\label{tab:tab1_FIQ}
\begin{tabular}{ccccccccc}
\hline
\multicolumn{1}{c|}{} & \multicolumn{2}{c|}{Dresses} & \multicolumn{2}{c|}{Shirts} & \multicolumn{2}{c|}{Tops\&Tees} & \multicolumn{2}{c}{Avg} \\ \cline{2-9} 
\multicolumn{1}{c|}{\multirow{-2}{*}{Method}} & R@10 & \multicolumn{1}{c|}{R@50} & R@10 & \multicolumn{1}{c|}{R@50} & R@10 & \multicolumn{1}{c|}{R@50} & R@10 & R@50 \\ \hline
\multicolumn{9}{c}{\cellcolor[HTML]{EFEFEF}\textit{Traditional Model-Based Methods}} \\
\multicolumn{1}{c|}{TIRG \cite{vo2019composing} $_{(CVPR'19)}$ }& 14.87 & \multicolumn{1}{c|}{34.66} & 18.26 & \multicolumn{1}{c|}{37.89} & 19.08 & \multicolumn{1}{c|}{39.62} & 17.40 & 37.39 \\
\multicolumn{1}{c|}{CIRPLANT \cite{liu2021image} $_{(ICCV'19)}$} & 17.45 & \multicolumn{1}{c|}{40.41} & 17.53 & \multicolumn{1}{c|}{38.81} & 21.64 & \multicolumn{1}{c|}{45.38} & 18.87 & 41.53 \\
\multicolumn{1}{c|}{ARTEMIS \cite{delmas2022artemis} $_{(ICLR'22)}$} & 27.16 & \multicolumn{1}{c|}{52.40} & 21.78 & \multicolumn{1}{c|}{43.64} & 29.20 & \multicolumn{1}{c|}{54.83} & 26.05 & 50.29 \\
\multicolumn{9}{c}{\cellcolor[HTML]{EFEFEF}\textit{VLP Model-Based Methods}} \\
\multicolumn{1}{c|}{CLIP4CIR \cite{baldrati2023composed} $_{(TOMM'23)}$} & 33.81 & \multicolumn{1}{c|}{59.40} & 39.99 & \multicolumn{1}{c|}{60.45} & 41.41 & \multicolumn{1}{c|}{65.37} & 38.40 & 61.74 \\
\multicolumn{1}{c|}{TGCIR \cite{wen2023target} $_{(ACM\ MM'23)}$} & 45.22 & \multicolumn{1}{c|}{69.66} & 52.60 & \multicolumn{1}{c|}{72.52} & 56.14 & \multicolumn{1}{c|}{77.10} & 51.32 & 73.09 \\
\multicolumn{1}{c|}{FashionERN \cite{chen2024fashionern} $_{(AAAI'24)}$} & 50.32 & \multicolumn{1}{c|}{71.29} & 50.15 & \multicolumn{1}{c|}{70.36} & 56.40 & \multicolumn{1}{c|}{77.21} & 52.29 & 72.95 \\
\multicolumn{1}{c|}{SADN \cite{wang2024semantic} $_{(ACM\ MM'24)}$} & 40.01 & \multicolumn{1}{c|}{65.10} & 43.67 & \multicolumn{1}{c|}{66.05} & 48.04 & \multicolumn{1}{c|}{70.93} & 43.91 & 67.36 \\
\multicolumn{1}{c|}{DQU-CIR \cite{wen2024simple} $_{(SIGIR'24)}$} & \underline{57.63} & \multicolumn{1}{c|}{\underline{78.56}}  & \underline{62.14} & \multicolumn{1}{c|}{\underline{80.38}} & \underline{66.15} & \multicolumn{1}{c|}{\underline{85.73}} & \underline{61.97} & \underline{81.56} \\
% \multicolumn{1}{c|}{DWC \cite{huang2024dynamic} $_{(AAAI'24)}$} & 32.67 & \multicolumn{1}{c|}{57.96} & 35.53 & \multicolumn{1}{c|}{60.11} & 40.13 & \multicolumn{1}{c|}{66.09} & 36.11 & 61.39 \\
\multicolumn{1}{c|}{CaLa \cite{jiang2024cala} $_{(SIGIR'24)}$} & 42.38 & \multicolumn{1}{c|}{66.08} & 46.76 & \multicolumn{1}{c|}{68.16} & 50.93 & \multicolumn{1}{c|}{73.42} & 46.69 & 69.22 \\
\multicolumn{1}{c|}{SSN \cite{yang2024decomposing} $_{(AAAI'24)}$} & 34.36 & \multicolumn{1}{c|}{60.78} & 38.13 & \multicolumn{1}{c|}{61.83} & 44.26 & \multicolumn{1}{c|}{69.05} & 38.92 & 63.89 \\
\multicolumn{1}{c|}{CASE \cite{levy2024data} $_{(AAAI'24)}$} & 47.44 & \multicolumn{1}{c|}{69.36} & 48.48 & \multicolumn{1}{c|}{70.23} & 50.18 & \multicolumn{1}{c|}{72.24} & 48.70 & 70.61 \\
\multicolumn{1}{c|}{CoVRBLIP \cite{ventura2024covr} $_{(AAAI'24)}$} & 44.55 & \multicolumn{1}{c|}{69.03} & 48.43 & \multicolumn{1}{c|}{67.42} & 52.60 & \multicolumn{1}{c|}{74.31} & 48.53 & 70.25 \\
\multicolumn{1}{c|}{SPRC \cite{bai2023sentence} $_{(ICLR'24)}$} & 49.18 & \multicolumn{1}{c|}{72.43} & 55.64 & \multicolumn{1}{c|}{73.89} & 59.35 & \multicolumn{1}{c|}{78.58} & 54.72 & 74.97 \\
\multicolumn{1}{c|}{ENCODER \cite{li2025encoder} $_{(AAAI'25)}$} & 51.51 & \multicolumn{1}{c|}{76.95} & 54.86 & \multicolumn{1}{c|}{74.93} & 62.01 & \multicolumn{1}{c|}{80.88} & 56.13 & 77.59 \\
\multicolumn{1}{c|}{CIRLVLM \cite{sun2025leveraging} $_{(AAAI'25)}$} & 50.42 & \multicolumn{1}{c|}{73.57} & 58.59 & \multicolumn{1}{c|}{75.86} & 59.61 & \multicolumn{1}{c|}{78.99} & 56.21 & 76.14 \\
\multicolumn{1}{c|}{CCIN \cite{tian2025ccin} $_{(CVPR'25)}$} & 49.38 & \multicolumn{1}{c|}{72.58} & 55.93 & \multicolumn{1}{c|}{74.14} & 57.93 & \multicolumn{1}{c|}{77.56} & 54.41 & 74.76 \\ \hline
\multicolumn{1}{c|}{\textbf{MCoT-MVS$_{\rm (Ours)}$}} & \textbf{58.45} & \multicolumn{1}{c|}{\textbf{78.92}} & \textbf{63.24} & \multicolumn{1}{c|}{\textbf{81.15}} & \textbf{68.02} & \multicolumn{1}{c|}{\textbf{85.97}} & \textbf{63.24} & \textbf{82.01} \\ \hline
\end{tabular}
\end{table*}

\section{Experiments}
\subsection{Dataset and Experiment Settings}
\textbf{Dataset.}
To verify the effectiveness, we conduct extensive experiments on two standard composed image retrieval benchmarks, \textit{i.e.} CIRR~\cite{liu2021image} and FashionIQ \cite{wu2021fashion}.
CIRR is an open-world scene image retrieval dataset with 36,554 triplets from 21,552 NLVR2 images~\cite{suhr2018corpus}, split 8:1:1 for training, validation, and testing. Fashion-IQ contains 30,134 triplets from 77,684 fashion images, divided into Dress, Shirt, and Toptee categories. Fashion-IQ targets a specialised domain, whereas CIRR emphasises diverse object interactions in real-world scenes. Evaluations on both datasets better reflect the method’s real-world generalization ability.

\noindent\textbf{Evaluation metrics.}
For performance comparison, following prior works \cite{tian2025ccin}, we report average recall at rank K (Recall@K). For CIRR, we provide Recall@1, 5, 10, and 50, along with their mean. For FashionIQ, we report Recall@10 and 50 across the three categories and their mean to evaluate the model’s generalization within specific fashion domains.

\noindent\textbf{Implementation Details.}
We use a Qwen2.5-VL-32B-Instruct\footnote{Qwen2.5-VL-32B-Instruct: https://huggingface.co/spaces/Qwen/Qwen2.5-VL-32B-Instruct} with a chain-of-thought (CoT) prompting strategy to obtain reasoning content given the multimodal composed query, including the retained text, the deleted text and the potential target texts. We use concurrent MLLM CoT batch inference acceleration, with a parallel rate of more than 1 image per second. For visual feature extraction, we employ the pre-trained CLIP model.% (ViT-H/14). 
We employ Grounded SAM \cite{ren2024grounded} to segment the objects in reference images with the box prediction threshold set to 0.4 and the text matching threshold set to 0.3.   During inference, MCoT reasoning takes $\thicksim$0.85s per query by API, segmentation takes $\thicksim$0.3s with $\thicksim$5.5GB GPU memory, and retrieval (with PVRS, IVRS, and WHC) takes $\thicksim$0.11s with $\thicksim$6.4GB memory.  

For training on the CIRR dataset, all parameters are fine-tuned with a learning rate of $1 \times 10^{-6}$. In contrast, for the FashionIQ dataset, the CLIP parameters are optimized at a learning rate of $1 \times 10^{-6}$, while the remaining parameters are trained with a learning rate of $1 \times 10^{-5}$ to facilitate better convergence. Similar to DQU-CIR \cite{wen2024simple}, we also write reasoned target text in the reference image to improve small modifications of simple images with good performance improvement on FasionIQ verified by DQU-CIR. However, in complex CIRR scenes, the improvement was not significant (only 0.4\% in averaged Recalls) after verification.
We adopt a fixed batch size of $B = 16$ for all experiments. The temperature factor $\tau$ in Eqn.~\eqref{eq:loss} is set to $0.01$ for CIRR and $0.1$ for FashionIQ. All experiments are implemented using PyTorch and conducted on a single NVIDIA A800 GPU. To ensure reproducibility, the random seed is set to 124 throughout all experiments.

\begin{table*}[t]
%\scriptsize
%\vspace{-0.5em}
\begin{center}
% \fontsize{10}{10.5}\selectfont
\fontsize{9}{10}\selectfont
\renewcommand\tabcolsep{5pt}
\caption{Ablation study for MCoT-MVS on CIRR. ``M-T" and ``T-T" indicate the modification texts and the reasoned target texts. }
\vspace{-1em}
\begin{tabular}{c|ccc|cc|cc|cccc|c}
\hline

\hline
\multirow{2}{*}{NO. } & \multicolumn{7}{c|}{Variant} & \multicolumn{5}{c}{Performance}    \\ \cline{2-13}
 & \multirow{1}{*}{WHC} & \multirow{1}{*}{Sum} & \multicolumn{1}{c|}{Combiner}  & \multirow{1}{*}{M-T} & \multirow{1}{*}{T-T} & PVRS & IVRS & Recall@1 & Recall@5  & Recall@10 & Recall@50 &  Mean  \\ \hline
1& \cellcolor{gray!10}$\surd$ & \cellcolor{gray!10}- & \cellcolor{gray!10}-  & \cellcolor{gray!10}$\surd$   & \cellcolor{gray!10}$\surd$ & \cellcolor{gray!10}$\surd$ & \cellcolor{gray!10}$\surd$   & \cellcolor{gray!10}\textbf{55.33} & 
  \cellcolor{gray!10}\textbf{84.75} & 
 \cellcolor{gray!10}\textbf{91.45} & 
 \cellcolor{gray!10}\textbf{98.55} &  
 \cellcolor{gray!10}\textbf{82.52}  \\% \cline{1-1} \cline{5-11}
 2& -& $\surd$& -  & $\surd$  & $\surd$ &$\surd$ &$\surd$   & 52.02 &  83.01 & 90.80 &  98.36 &  81.05 \\ \cdashline{2-13}
 % 3& -& -   & $\surd$   & $\surd$  &$\surd$ &$\surd$   & xx &  xx & xx &  xx &  xx \\  \cdashline{2-12}
  3& -  & -   & $\surd$ &- & $\surd$  &$\surd$ &$\surd$   & 45.00 & 76.48  & 85.35 & 96.87 &  75.93  \\ 
 4& -  & -   & $\surd$ &$\surd$ & -  &$\surd$ &$\surd$   & 54.50 & 83.69& 91.28 &  98.56 &  81.98 \\ \cdashline{2-13} %\cdashline{5-6}
 5&   & -   & $\surd$ &$\surd$ & -  &$\surd$ &-   & 53.49 & 84.07 & 91.21 & 98.55&  81.83 \\ 
 6& - & - & $\surd$ &$\surd$ & -  & -  &$\surd$   & 53.91 & 83.81 & 90.87 & 98.53 &  81.78 \\   \cdashline{2-13}
 7& - & - & $\surd$&$\surd$  & -  & -  &-   & 51.66   & 82.61  & 90.21 & 98.60 &  80.77 \\
 8& - & $\surd$& - &$\surd$  & -  & -  &-   & 49.61  & 81.61 & 89.74 & 98.55 &  79.87 \\ \hline 
 
 \hline
\end{tabular}
\label{tab:tab3_ab}
\end{center}
\vspace{-1em}
\end{table*}

\subsection{State-of-the-art Comparisons}
We perform extensive experiments to compare our MCoT-MVS with mainstream CIR studies and the latest state-of-the-art methods on two widely-used CIR benchmarks, i.e. CIRR and FashionIQ, in Table \ref{tab:tab1_CIRR} and Table \ref{tab:tab1_FIQ}, respectively.  %The best and second-best results are bold and underlined.

\noindent\textbf{Quantitative Comparison on CIRR:}
% We compare our proposed MCoT-MVS with its counterpart in Table \ref{tab:table1}.
On the CIRR dataset, which features complex textual modifications and semantically rich images, our MCoT-MVS consistently surpasses mainstream approaches with impressive margins across all evaluation metrics in Table \ref{tab:tab1_CIRR}. 
Compared to the latest state-of-the-art CCIN~\cite{tian2025ccin}, which also employs a pre-trained LLM to reason about user intent by reasoning image-caption-derived queries into conflict and preserved sets, MCoT-MVS achieves a 1.92\% gain in Recall@1 and further improves performance at higher cutoffs (e.g., Recall@5 and Recall@50), leading to a 0.86\% average improvement across all metrics. These consistent gains demonstrate that explicitly decomposing and reasoning about user intent allows our MCoT-MVS to better capture multi-level visual references guided by the textual modification, thereby delivering more accurate and robust retrieval results on the challenging benchmark.

\noindent\textbf{Quantitative Comparison on FashionIQ:} Table \ref{tab:tab1_FIQ} also lists the result of competing methods on the FashionIQ dataset. Our MCoT-MVS also achieves best retrieval performance on all eight evaluation metrics in three categories, including ``Dresses", ``Shirts", ``Tops\&Tees", and corresponding average scores.
In particular, compared to the latest method CIR-LVLM \cite{sun2025leveraging}, which fine-tunes an MLLM to refine user intent instructions, MCoT-MVS yields notable gains of 8.1\%, 4.6\%, and 8.4\% in Recall@10 for the ``Dresses," ``Shirts", and ``Tops\&Tees" categories, respectively.
Additionally, compared with the state-of-the-art DQU-CIR under the same data processing settings, our MCoT-MVS achieves improvements of 1.2\% and 0.4\% in average Recall@10 and Recall@50, respectively.

Overall, these results on two different-domain datasets demonstrate the powerful effectiveness of the proposed MCoT-MVS model with strong generalization capability.

\subsection{Ablation Studies}
We perform detailed ablation studies on the CIRR dataset to investigate the effectiveness of each component of our proposed MCoT-MVS. The detailed results are shown in Table \ref{tab:tab3_ab}.

\noindent\textbf{The Effectiveness of the Weighted Hierarchical Combination Module.}
We conduct different feature combinations to observe the performance of different multimodal query feature combinations and evaluate the superiority of our proposed Weighted Hierarchical Combination (WHC) module. 
% First, comparing baseline 7 and baseline 8 reveals that the combiner improves the retrieval performance by combining multimodal query features with weights. 
First, comparing Baseline 7 and Baseline 8 reveals that using a learnable combiner to fuse multimodal query features leads to notable performance gains. It suggests that composed queries inherently contain modality intent ambiguity, which benefits from adaptive weighting. %, proving that combined queries have the uncertainty of modality intent representation.
Second, by comparing full MCoT-MVS (No.1) with Variant NO.2, where the proposed WHC module is replaced by direct summation of four types of multimodal query representations, the retrieval performance substantially drops from  82.52\% to 81.05\% in average recall. These observations demonstrate that our WHC module not only effectively integrates the intent information from the modified text (e.g. appearance of new objects) but also effectively filters out potential errors from target reasoning, while preserving the effective selection of the reference image.

\noindent\textbf{Effects of the MLLM-based Reasoned Target Text.}
To evaluate the impact of potential target content reasoned from the multimodal composed query via MLLM, we first directly replace modification texts with the reasoned target texts to retrieve target images (indicated Variant No.3). This variant achieves an average retrieval performance of 75.93\%, demonstrating the validity of the reasoned content. %The main reason why its effect is not as good as modified text is that we believe that multimodal reasoning still has a certain degree of illusion, while modified text is more accurate, but the ability of target reasoning is still good. 
However, its effectiveness is still lower than that of using the original modification text, which we attribute to the fact that multimodal reasoning may introduce a certain degree of hallucination, while modification text generally provides more precise intent cues. Nevertheless, the results confirm that the reasoning ability is meaningful and beneficial.
By comparing the performance of our MCoT-MVS (No.1) with Variant 3 and Variant 4, it is evident that integrating the reasoned target content alongside the original modification text provides additional accurate and effective intent information for alignment with the target image. It also proves the effectiveness of our multimodal CoT reasoning.

\noindent\textbf{The Effectiveness of Multi-level Visual Reference Selection (PVRS and IVRS).} 
We construct extensive ablation studies, including experimental variants No.4 to No.7, to evaluate the impact of our proposed patch-level visual reference selection module (PVRS) and instance-level visual reference selection module. Specifically, when removing the PVRS module, the Recall@1 performance decreases absolutely by 0.59\% compared with variant No.4. It suggests that patch-level reference image de-noising guided by the reasoned retained and removed user intent plays a vital role in our model.
Similarly, when removing the IVRS module, it leads to an absolute 1.01\% drop in Recall@1. By independently modelling instances in the reference image, the key instances of user intent can be clarified sufficiently and effectively, and the user intent representation can be enhanced from the overall semantics and even the spatial shape of the reference image instances.
Furthermore, when both PVRS and IVRS are removed, the overall retrieval performance drops substantially from 81.98\% to 80.77\%, demonstrating that the two modules are complementary. These results strongly validate the effectiveness of our proposed multi-level visual reference selection mechanism in MCoT-MVS for improving the interpretation and clarification of multimodal composed queries in the CIR task.

\begin{figure*}[t] %%%%%%%%%%%%%%%%%fig2
	\centering
	% \vspace{0.3em}
	\includegraphics[width=0.9\linewidth]{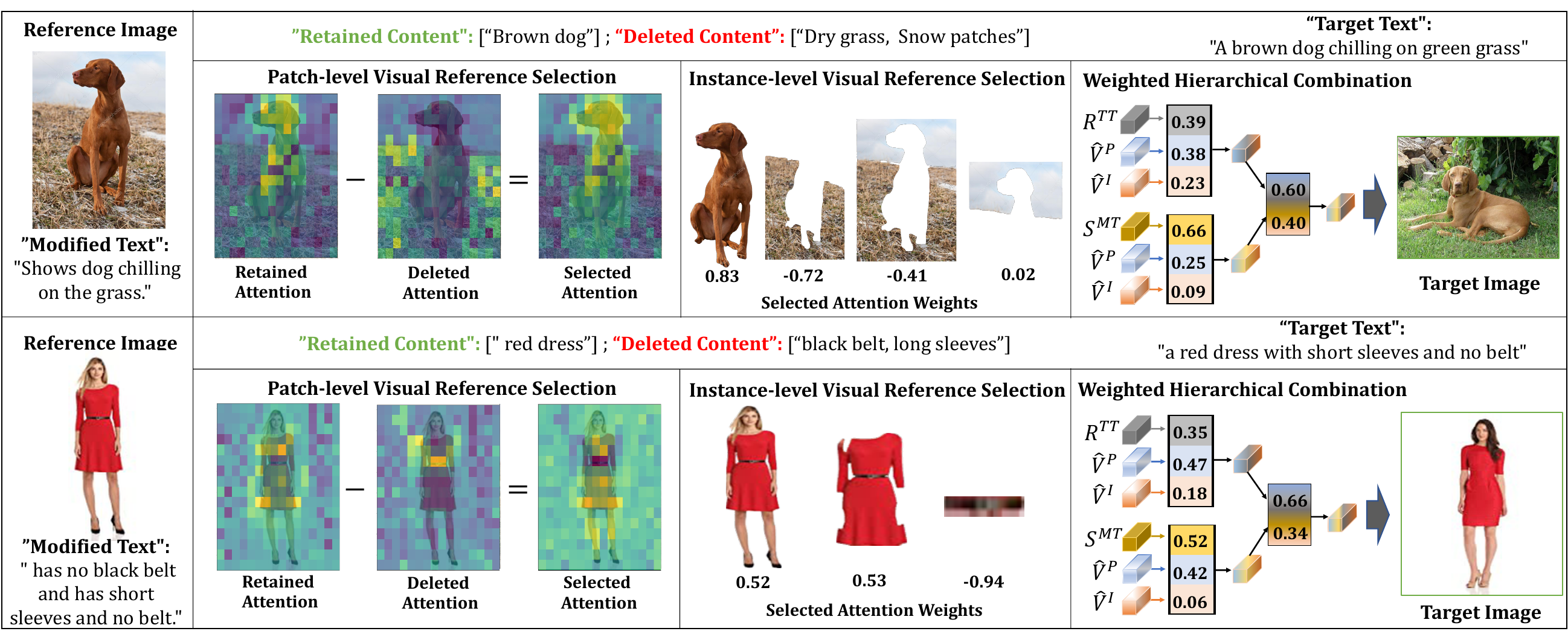}
	\vspace{-1em}
	\caption{Visualization of the learned attention weights from PVRS and IVRS modules guided by the inferred explicit retain or delete modification intents from the MLLM-based multimodal CoT reasoning, as well as the weight distributions of weighted hierarchical combination modules (best viewed in color).} 
	\label{fig:visual_attn_01}
	\vspace{-1em}
\end{figure*}   
\begin{table}[t]
% \begin{threeparttable}
\centering
\caption{Retrieval performance (\%) of MCoT-MVS with different reasoning patterns.}
\vspace{-1em}
\begin{tabular}{lccccc}
\toprule
Model & R@1 & R@5 & R@10 & R@50 \\
\midrule
w/o Multi-step MCot  & 54.29  & 84.45 & {91.15} & \textbf{98.60} \\
w/ Multi-step MCot & \textbf{55.33}  & \textbf{84.75} & \textbf{91.45} & {98.55} \\
\bottomrule
\end{tabular}
\label{tab:MCot par}
% \end{threeparttable}
\end{table}
\begin{table}[t]
% \begin{threeparttable}
\centering
\vspace{-1em}
\caption{Retrieval performance (\%) of MCoT-MVS with different reasoning models.}
\vspace{-1em}
\begin{tabular}{lccccc}
\toprule
Model & R@1 & R@5 & R@10 & R@50 \\
\midrule
Llama-3.2-11B\footnotemark[1] & 54.31  & 84.26 & \textbf{91.78} & \textbf{98.69} \\
Qwen2.5-VL-7B\footnotemark[2] & 54.53 & 84.31 & 91.13 & 98.50 \\
Qwen2.5-VL-32B & \textbf{55.33}  & \textbf{84.75} & 91.45 & 98.55 \\
\bottomrule
\end{tabular}
\label{tab:MLLM}
% \end{threeparttable}
\vspace{-1em}
\end{table}
\footnotetext[1]{Llama-3.2-11B: https://huggingface.co/meta-llama/Llama-3.2-11B-Vision}
\footnotetext[2]{Qwen2.5-VL-7B: https://huggingface.co/Qwen/Qwen2.5-VL-7B-Instruct}
\subsection{The Impact of Chain-of-Thought Reasoning}
We conduct the comparison experiments in Table \ref{tab:MCot par} to validate the importance of multi-step multimodal chain-of-thought (MCoT) reasoning. 
In particular, when reasoning about the explicit use intent without the multi-step CoT, i.e., directly letting the MLLM predict the ``retain/delete/target” pattern, the retrieval performance decreases, especially on the higher top-ranking. The main reason for this is that multi-step chain-of-thought reasoning can yield a more precise decomposition of user intent. Although performance without a multi-step MCoT is not optimal, the displayed user intent decomposition still outperforms the baseline (No. 7 in Table \ref{tab:tab3_ab}) without it. This further demonstrates the effectiveness of displaying the decomposed user intent and making reference visual selections.

\subsection{The Impact of Reasoning Abilities}
To further examine the sensitivity of our proposed MCoT-MVS framework to the reasoning abilities of different MLLMs, we conduct experiments under three representative multimodal reasoning models: Llama-3.2-11B-Vision-Instruct, Qwen2.5-VL-7B, and Qwen2.5-VL-32B.

As shown in Table~\ref{tab:MLLM}, MCoT-MVS consistently achieves strong retrieval performance across all reasoning models. Although the larger Qwen2.5-VL-32B achieves the best overall accuracy (R@1: 55.325), the smaller Llama-3.2-11B and Qwen2.5-VL-7B still maintain highly competitive results. This demonstrates two key findings: (i) MCoT-MVS has a certain sensitivity to improvements in reasoning power, as more powerful reasoning models can lead to retrieval improvements by better understanding user intent; and (ii) MCoT-MVS is robust to weaker reasoning ability, as performance does not collapse even when reasoning power decreases, maintaining strong performance compared to other models. These results demonstrate that MCoT-MVS effectively constrains reasoning output and maintains reliability across varying model capacities.

\subsection{Qualitative Analysis}
To better understand the contribution of each component in MCoT-MVS, we visualize the attention weights from PVRS and IVRS modules in Figure \ref{fig:visual_attn_01}, guided by multimodal CoT reasoning.  Specifically, the PVRS module performs semantic denoising at the patch-level by enhancing areas aligned with the preserved intent and attenuating those related to modified or deleted content, thus enabling more fine-grained visual filtering. Meanwhile, the IVRS module operates at the high-level instance-level to capture key target instances mentioned in the user’s intent by modeling their appearance features while discarding distracting instances. These attention maps clearly illustrate the complementary effects of the two modules at different granularity levels, leading to enhanced visual intent representation and ultimately improved query interpretation and retrieval performance in CIR tasks.

Additionally, we also visualize the learned weights from WHC module. In both examples, the reasoned target text receives higher weights than the original modification text, better capturing user intent and validating the effectiveness of our multimodal CoT reasoning.  For visual references, the patch-level selected features are assigned greater weights due to their finer granularity, while the instance-level features also provide meaningful complementary cues. These results demonstrate that WHC can effectively allocate fusion weights across multimodal inputs, enhancing intent representation and improving CIR performance.

\section{Conclusion and Future Work}
In this paper, we propose MCoT-MVS, a novel method for Composed Image Retrieval (CIR) that explicitly selects multi-level reference visual features under the guidance of multimodal Chain-of-Thought (CoT) reasoning. By disentangling and reconstructing fine-grained and high-level visual cues based on explicit retain or delete modification intents, our approach significantly enhances visual intent representation. Furthermore, a weighted hierarchical attention combination mechanism (WHC) effectively integrates the modified text, reasoned target content, and multi-level visual selected references, enabling the model to suppress irrelevant information and preserve a coherent fusion aligned with the target image. Extensive quantitative evaluations on two standard CIR benchmarks, CIRR and FashionIQ, demonstrate that MCoT-MVS achieves state-of-the-art performance across multiple retrieval metrics.

On one hand, while the target content of multimodal CoT reasoning is available, its effectiveness still needs to be improved. On the other hand, multi-step MCoT reasoning has a certain impact on reasoning efficiency, and direct reasoning also provides good performance for user intent. Therefore, in future work, we will further explore hybrid reasoning paradigms to improve the accuracy of target reasoning while further enhancing the efficiency of the reasoning process.

\begin{acks}
This research was (partially) supported by the Natural Science Foundation of China (62202271, 62472261, 62372275, 62272274, T2293773), the National Key R\&D Program of China with grant No. 2024YFC3307300 and No. 2022YFC3303004, the Provincial Key R\&D Program of Shandong Province with grant No. 2024CXGC010108, the Natural Science Foundation of Shandong Province with grant No. ZR2024QF203, the Technology Innovation Guidance Program of Shandong Province with grant No. YDZX2024088. All content represents the opinion of the authors, which is not necessarily shared or endorsed by their respective employers and/or sponsors.
\end{acks}

\bibliographystyle{ACM-Reference-Format}
\balance
\bibliography{sample-base}

@String{Computing = "Computing" }

@String{Computer = "{IEEE} Computer" }

@String{Springer = "Springer-Verlag" }

@article{fu2025efficient,
  title={Efficient and effective adaptation of multimodal foundation models in sequential recommendation},
  author={Fu, Junchen and Ge, Xuri and Xin, Xin and Karatzoglou, Alexandros and Arapakis, Ioannis and Zheng, Kaiwen and Ni, Yongxin and Joemon, Joemon M Jose},
  journal={IEEE Transactions on Knowledge and Data Engineering},
  year={2025},
  publisher={IEEE}
}

@inproceedings{chen2020image,
  title={Image search with text feedback by visiolinguistic attention learning},
  author={Chen, Yanbei and Gong, Shaogang and Bazzani, Loris},
  booktitle={Proceedings of the IEEE/CVF Conference on Computer Vision and Pattern Recognition},
  pages={3001--3011},
  year={2020}
}

@article{xu2023multi,
  title={Multi-modal transformer with global-local alignment for composed query image retrieval},
  author={Xu, Yahui and Bin, Yi and Wei, Jiwei and Yang, Yang and Wang, Guoqing and Shen, Heng Tao},
  journal={IEEE Transactions on Multimedia},
  volume={25},
  pages={8346--8357},
  year={2023},
  publisher={IEEE}
}

@inproceedings{wen2021comprehensive,
  title={Comprehensive linguistic-visual composition network for image retrieval},
  author={Wen, Haokun and Song, Xuemeng and Yang, Xin and Zhan, Yibing and Nie, Liqiang},
  booktitle={Proceedings of the 44th International ACM SIGIR Conference on Research and Development in Information Retrieval},
  pages={1369--1378},
  year={2021}
}

@inproceedings{tian2023fashion,
  title={Fashion image retrieval with text feedback by additive attention compositional learning},
  author={Tian, Yuxin and Newsam, Shawn and Boakye, Kofi},
  booktitle={Proceedings of the IEEE/CVF winter conference on applications of computer vision},
  pages={1011--1021},
  year={2023}
}

@inproceedings{wu2021fashion,
  title={Fashion iq: A new dataset towards retrieving images by natural language feedback},
  author={Wu, Hui and Gao, Yupeng and Guo, Xiaoxiao and Al-Halah, Ziad and Rennie, Steven and Grauman, Kristen and Feris, Rogerio},
  booktitle={The IEEE Conference on Computer Vision and Pattern Recognition},
  pages={11307--11317},
  year={2021}
}

@inproceedings{liu2021image,
  title={Image retrieval on real-life images with pre-trained vision-and-language models},
  author={Liu, Zheyuan and Rodriguez-Opazo, Cristian and Teney, Damien and Gould, Stephen},
  booktitle={The IEEE International Conference on Computer Vision},
  pages={2125--2134},
  year={2021}
}

@article{suhr2018corpus,
  title={A corpus for reasoning about natural language grounded in photographs},
  author={Suhr, Alane and Zhou, Stephanie and Zhang, Ally and Zhang, Iris and Bai, Huajun and Artzi, Yoav},
  journal={In Proceedings of the 57th Annual Meeting of the Association for Computational Linguistics},
  year={2019}
}

@inproceedings{tian2025ccin,
  title={CCIN: Compositional Conflict Identification and Neutralization for Composed Image Retrieval},
  author={Tian, Likai and Zhao, Jian and Hu, Zechao and Yang, Zhengwei and Li, Hao and Jin, Lei and Wang, Zheng and Li, Xuelong},
  booktitle={Proceedings of the Computer Vision and Pattern Recognition Conference},
  pages={3974--3983},
  year={2025}
}

@inproceedings{vo2019composing,
  title={Composing text and image for image retrieval-an empirical odyssey},
  author={Vo, Nam and Jiang, Lu and Sun, Chen and Murphy, Kevin and Li, Li-Jia and Fei-Fei, Li and Hays, James},
  booktitle={Proceedings of the IEEE/CVF conference on computer vision and pattern recognition},
  pages={6439--6448},
  year={2019}
}

@article{sun2025cotmr,
  title={CoTMR: Chain-of-Thought Multi-Scale Reasoning for Training-Free Zero-Shot Composed Image Retrieval},
  author={Sun, Zelong and Jing, Dong and Lu, Zhiwu},
  journal={arXiv preprint arXiv:2502.20826},
  year={2025}
}

@inproceedings{radford2021learning,
  title={Learning transferable visual models from natural language supervision},
  author={Radford, Alec and Kim, Jong Wook and Hallacy, Chris and Ramesh, Aditya and Goh, Gabriel and Agarwal, Sandhini and Sastry, Girish and Askell, Amanda and Mishkin, Pamela and Clark, Jack and others},
  booktitle={International conference on machine learning},
  pages={8748--8763},
  year={2021},
  organization={PmLR}
}

@article{ren2024grounded,
  title={Grounded sam: Assembling open-world models for diverse visual tasks},
  author={Ren, Tianhe and Liu, Shilong and Zeng, Ailing and Lin, Jing and Li, Kunchang and Cao, He and Chen, Jiayu and Huang, Xinyu and Chen, Yukang and Yan, Feng and others},
  journal={International Conference on Computer Vision (ICCV) Demo Paper},
  year={2024}
}

@article{delmas2022artemis,
  title={Artemis: Attention-based retrieval with text-explicit matching and implicit similarity},
  author={Delmas, Ginger and de Rezende, Rafael Sampaio and Csurka, Gabriela and Larlus, Diane},
  journal={International Conference on Learning Representations},
  year={2022}
}

@article{baldrati2023composed,
  title={Composed image retrieval using contrastive learning and task-oriented clip-based features},
  author={Baldrati, Alberto and Bertini, Marco and Uricchio, Tiberio and Del Bimbo, Alberto},
  journal={ACM Transactions on Multimedia Computing, Communications and Applications},
  volume={20},
  number={3},
  pages={1--24},
  year={2023},
  publisher={ACM New York, NY}
}

@inproceedings{chen2020learning,
  title={Learning joint visual semantic matching embeddings for language-guided retrieval},
  author={Chen, Yanbei and Bazzani, Loris},
  booktitle={European Conference on Computer Vision},
  pages={136--152},
  year={2020},
  organization={Springer}
}

@inproceedings{wen2023target,
  title={Target-guided composed image retrieval},
  author={Wen, Haokun and Zhang, Xian and Song, Xuemeng and Wei, Yinwei and Nie, Liqiang},
  booktitle={Proceedings of the 31st ACM international conference on multimedia},
  pages={915--923},
  year={2023}
}

@inproceedings{chen2024fashionern,
  title={FashionERN: enhance-and-refine network for composed fashion image retrieval},
  author={Chen, Yanzhe and Zhong, Huasong and He, Xiangteng and Peng, Yuxin and Zhou, Jiahuan and Cheng, Lele},
  booktitle={Proceedings of the AAAI Conference on Artificial Intelligence},
  volume={38},
  number={2},
  pages={1228--1236},
  year={2024}
}

@inproceedings{wang2024semantic,
  title={Semantic Distillation from Neighborhood for Composed Image Retrieval},
  author={Wang, Yifan and Huang, Wuliang and Li, Lei and Yuan, Chun},
  booktitle={Proceedings of the 32nd ACM International Conference on Multimedia},
  pages={5575--5583},
  year={2024}
}

@inproceedings{wen2024simple,
  title={Simple but effective raw-data level multimodal fusion for composed image retrieval},
  author={Wen, Haokun and Song, Xuemeng and Chen, Xiaolin and Wei, Yinwei and Nie, Liqiang and Chua, Tat-Seng},
  booktitle={Proceedings of the 47th International ACM SIGIR conference on research and development in information retrieval},
  pages={229--239},
  year={2024}
}

@inproceedings{jiang2024cala,
  title={Cala: Complementary association learning for augmenting comoposed image retrieval},
  author={Jiang, Xintong and Wang, Yaxiong and Li, Mengjian and Wu, Yujiao and Hu, Bingwen and Qian, Xueming},
  booktitle={Proceedings of the 47th International ACM SIGIR Conference on Research and Development in Information Retrieval},
  pages={2177--2187},
  year={2024}
}

@inproceedings{yang2024decomposing,
  title={Decomposing semantic shifts for composed image retrieval},
  author={Yang, Xingyu and Liu, Daqing and Zhang, Heng and Luo, Yong and Wang, Chaoyue and Zhang, Jing},
  booktitle={Proceedings of the AAAI Conference on Artificial Intelligence},
  volume={38},
  number={7},
  pages={6576--6584},
  year={2024}
}

@inproceedings{levy2024data,
  title={Data roaming and quality assessment for composed image retrieval},
  author={Levy, Matan and Ben-Ari, Rami and Darshan, Nir and Lischinski, Dani},
  booktitle={Proceedings of the AAAI conference on artificial intelligence},
  volume={38},
  number={4},
  pages={2991--2999},
  year={2024}
}

@inproceedings{ventura2024covr,
  title={Covr: Learning composed video retrieval from web video captions},
  author={Ventura, Lucas and Yang, Antoine and Schmid, Cordelia and Varol, G{\"u}l},
  booktitle={Proceedings of the AAAI Conference on Artificial Intelligence},
  volume={38},
  number={6},
  pages={5270--5279},
  year={2024}
}

@article{bai2023sentence,
  title={Sentence-level prompts benefit composed image retrieval},
  author={Bai, Yang and Xu, Xinxing and Liu, Yong and Khan, Salman and Khan, Fahad and Zuo, Wangmeng and Goh, Rick Siow Mong and Feng, Chun-Mei},
  journal={International Conference on Learning Representations},
  year={2024}
}

@inproceedings{kim2021dual,
  title={Dual compositional learning in interactive image retrieval},
  author={Kim, Jongseok and Yu, Youngjae and Kim, Hoeseong and Kim, Gunhee},
  booktitle={Proceedings of the AAAI Conference on Artificial Intelligence},
  volume={35},
  number={2},
  pages={1771--1779},
  year={2021}
}

@inproceedings{sun2025leveraging,
  title={Leveraging large vision-language model as user intent-aware encoder for composed image retrieval},
  author={Sun, Zelong and Jing, Dong and Yang, Guoxing and Fei, Nanyi and Lu, Zhiwu},
  booktitle={Proceedings of the AAAI Conference on Artificial Intelligence},
  volume={39},
  number={7},
  pages={7149--7157},
  year={2025}
}

@inproceedings{li2025encoder,
  title={Encoder: Entity mining and modification relation binding for composed image retrieval},
  author={Li, Zixu and Chen, Zhiwei and Wen, Haokun and Fu, Zhiheng and Hu, Yupeng and Guan, Weili},
  booktitle={Proceedings of the AAAI Conference on Artificial Intelligence},
  volume={39},
  number={5},
  pages={5101--5109},
  year={2025}
}

@article{karthik2023vision,
  title={Vision-by-language for training-free compositional image retrieval},
  author={Karthik, Shyamgopal and Roth, Karsten and Mancini, Massimiliano and Akata, Zeynep},
  journal={In Proceedings of the IEEE conference on computer vision and pattern recognition},
  year={2023}
}

@inproceedings{xie2020modeling,
  title={Modeling user behavior for vertical search: images, apps and products},
  author={Xie, Xiaohui and Mao, Jiaxin and Liu, Yiqun and de Rijke, Maarten},
  booktitle={Proceedings of the 43rd International ACM SIGIR Conference on Research and Development in Information Retrieval},
  pages={2440--2443},
  year={2020}
}

@article{wu2024personalized,
  title={Personalized prompt for sequential recommendation},
  author={Wu, Yiqing and Xie, Ruobing and Zhu, Yongchun and Zhuang, Fuzhen and Zhang, Xu and Lin, Leyu and He, Qing},
  journal={IEEE Transactions on Knowledge and Data Engineering},
  volume={36},
  number={7},
  pages={3376--3389},
  year={2024},
  publisher={IEEE}
}

@inproceedings{hosseinzadeh2020composed,
  title={Composed query image retrieval using locally bounded features},
  author={Hosseinzadeh, Mehrdad and Wang, Yang},
  booktitle={Proceedings of the IEEE/CVF conference on computer vision and pattern recognition},
  pages={3596--3605},
  year={2020}
}

@inproceedings{luo2025imagescope,
  title={ImageScope: Unifying Language-Guided Image Retrieval via Large Multimodal Model Collective Reasoning},
  author={Luo, Pengfei and Zhou, Jingbo and Xu, Tong and Xia, Yuan and Xu, Linli and Chen, Enhong},
  booktitle={Proceedings of the ACM on Web Conference 2025},
  pages={1666--1682},
  year={2025}
}

@article{gudivada1995content,
  title={Content based image retrieval systems},
  author={Gudivada, Venkat N and Raghavan, Vijay V},
  journal={Computer},
  volume={28},
  number={9},
  pages={18--22},
  year={1995},
  publisher={IEEE}
}

@article{guan2022partially,
  title={Partially supervised compatibility modeling},
  author={Guan, Weili and Wen, Haokun and Song, Xuemeng and Wang, Chun and Yeh, Chung-Hsing and Chang, Xiaojun and Nie, Liqiang},
  journal={IEEE Transactions on Image Processing},
  volume={31},
  pages={4733--4745},
  year={2022},
  publisher={IEEE}
}

\end{document}